%% file: sympa.tex
\documentclass{article}

\PassOptionsToPackage{numbers, compress}{natbib}

\usepackage[final]{neurips_2020}
\usepackage{hyperref}       
\hypersetup{
    pdfsubject={Hermitian Symmetric Spaces for Graph Embeddings},
    pdftitle={Hermitian Symmetric Spaces for Graph Embeddings},
    pdfauthor={Federico Lopez},
    pdfkeywords={hermitian symmetric spaces, graph embeddings, siegel space, hyperbolic geometry, upper half space, bounded domain model, matrix models, deep learning, riemannian manifold learning, non-euclidean}
}

\usepackage[utf8]{inputenc} 
\usepackage[T1]{fontenc}    
\usepackage{hyperref}       
\usepackage{url}            
\usepackage{booktabs}       
\usepackage{amsfonts}       
\usepackage{nicefrac}       
\usepackage{microtype}      

\usepackage{todonotes}
\usepackage{amsmath}
\usepackage{amssymb}
\usepackage{adjustbox}
\usepackage{subfig}			
\usepackage{graphicx}
\graphicspath{{./img/}}
\usepackage{algorithm,algorithmic}
\usepackage{caption}
\usepackage{wrapfig}

\newcommand{\Id}{{\rm Id}}
\newcommand{\R}{\mathbb R}
\newcommand{\Sym}{{\rm Sym}}
\newcommand{\C}{\mathbb C}
\newcommand{\Sp}{{\rm Sp}}
\newcommand{\calB}{\mathcal B}
\newcommand{\SL}{{\rm SL}}
\newcommand{\tr}{{\rm tr}}
\newcommand{\calS}{\mathcal S}
\newcommand{\bpm}{\begin{pmatrix}}
\newcommand{\epm}{\end{pmatrix}}
\newcommand{\bsm}{\left(\begin{smallmatrix}}
\newcommand{\esm}{\end{smallmatrix}\right)}                   
\newcommand{\ov}{\overline}

\title{Hermitian Symmetric Spaces for Graph Embeddings}

\author{%
  Federico L\'opez \\
  Heidelberg Institute for Theoretical Studies\\
  AIPHES\\
  \texttt{federico.lopez@h-its.org} \\
  \And
  Beatrice Pozzetti \\
  Heidelberg University\\
  \texttt{pozzetti@mathi.uni-heidelberg.de} \\
  \AND
  Steve Trettel\\
  Stanford University \\
  \texttt{stevejtrettel@gmail.com} \\
  \And
  Anna Wienhard \\
  Heidelberg University \\
  \texttt{wienhard@mathi.uni-heidelberg.de} \\
}

\begin{document}

\maketitle

\input{AbstractIntro}

\input{SymmSpaces}

\input{graphs}
\input{Experiments}

\input{Conclusions}

\section*{Broader Impact}

Graphs are ubiquitous, as data on a wide variety of domains can be modeled through them. Thus, learning faithful continuous representations of graphs is a fundamental step for several machine learning methods. We believe that our approach based on Riemannian manifolds such as Hermitian symmetric spaces should not raise any ethical considerations, to the extent that the input data for the model is unbiased. 
As such, we hope only a positive impact can emerge from our work, in the areas of social, biological, financial, or network data.


\begin{ack}
This work has been supported by the German Research Foundation (DFG) as part of the Research Training Group Adaptive Preparation of Information from Heterogeneous Sources (AIPHES) under grant No. GRK 1994/1, as well as under Germany’s Excellence Strategy EXC-2181/1 - 390900948 (the Heidelberg STRUCTURES Cluster of Excellence), and by the Klaus Tschira Foundation, Heidelberg, Germany. 
\end{ack}

\bibliographystyle{plainnat}
\small
\bibliography{sympa}

\newpage
\appendix
\include{appendix}

\end{document}

%% file: AbstractIntro.tex
\begin{abstract}



Learning faithful graph representations as sets of vertex embeddings has become a fundamental intermediary step in a wide range of machine learning applications.
The quality of the embeddings is usually determined by how well the geometry of the target space matches the structure of the data. 
In this work we learn continuous representations of graphs in spaces of symmetric matrices over $\C$. 
These spaces offer a rich geometry that simultaneously admits hyperbolic and Euclidean subspaces, and are amenable to analysis and explicit computations. 
We implement an efficient method to learn embeddings and compute distances, and develop the tools to operate with such spaces. The proposed models are able to automatically adapt to very dissimilar arrangements without any apriori estimates of graph features. 
On various datasets with very diverse structural properties and reconstruction measures our model ties the results of competitive baselines for geometrically pure graphs and outperforms them for graphs with mixed geometric features, showcasing the versatility of our approach.
\end{abstract}

\section{Introduction}
The goal of representation learning  is to embed real-world data, frequently modeled on a graph, into an ambient space. The structure of this embedding space can then be used to analyze and perform tasks on the discrete graph. 
The predominant approach has been to embed discrete structures in an Euclidean space. Nonetheless, data in many domains  including social \cite{lazer09lifeinnet, verbeek14social}, sensor \cite{gao2012sensor}, gene \cite{Davidson1669}, protein  molecular \cite{gainza2020molecules} and complex \cite{krioukov2010hypernetworks} networks or the Internet \cite{boguna2010internet} exhibit non-Euclidean features, making embeddings into Riemannian manifolds with a richer structure necessary. 
For this reason, embeddings into hyperbolic \cite{krioukov2009tempNets, Chamberlain2017, nickel2017poincare, nickel2018lorentz, deSa18tradeoffs} and spherical spaces \cite{wilson2014sphere, defferrard2019deepsphere} have been developed. 
Recent work proposes to combine different curvatures through several layers \cite{bachmann2020ccgcn, chami2019hgcnn}, or to enrich the geometry by considering Cartesian products of spaces \cite{gu2019lmixedCurvature, skopek2020mixedva, tifrea2018poincareGlove}.
However, their ability to model complex patterns is inherently bounded by the geometric properties of the target embedding space, which has to be picked beforehand. 
The choice of a metric space where to embed the data can be understood as selecting an inductive bias \cite{tifrea2018poincareGlove}, which does not always adapt to all parts of the graphs.

In this work, we learn representations of graphs in spaces of symmetric matrices with coefficients in the complex numbers $\C$. 
The matrix models are able to automatically adapt to very dissimilar patterns without any apriori estimates of the curvature or other features of the graph. 
By leveraging their rich geometry, which contains Euclidean subspaces, hyperbolic subspaces and products thereof, our model does not require a metric space selection or curvature calibration unlike previous work. 
The Hermitian symmetric spaces of non-constant sectional curvature that we introduce act as {\bf matrix models of the hyperbolic plane}. They are generalizations of the Poincar\'e disc and the upper half-plane model in spaces of matrices over $\C$. This offers several advantages over previous applications of positive definite matrix spaces \cite{cruceru20matrixGraph}, and provides practitioners with new analytical tools. By combining these tools with the Takagi factorization and the Cayley transform \cite{cayley1846transform}, we achieve a tractable and automatic-differentiable algorithm to compute distance in the complex matrix spaces. 

To evaluate the representation capacities of our models, we consider real and synthetic datasets, and graph reconstruction measures. We propose rooted products as a {\bf new class of benchmarking graphs}. They mix tree- and grid-like features at different levels and scales, making them a better approximation of real-world datasets.
We find that for synthetic graphs of pure geometry (i.e. trees, grids and their Cartesian products), Hermitian symmetric spaces are on par with the corresponding best geometric space. When we experiment with datasets of mixed geometric features, Hermitian spaces outperform all baselines. These results showcase the effectiveness and versatility of the proposed approach, particularly for graphs with varying and intricate structures.



Inspired by \citet{gu2019lmixedCurvature}, our work explores more flexible spaces of non-constant sectional curvature. 
Our approach fits in the area of Riemannian manifold learning \cite{bronstein2018geomdeeplearning}. It resembles the study of Grassmannian manifolds and the space of positive definite matrices \cite{cruceru20matrixGraph}. Moreover, we add a new layer of structure by considering embeddings into manifolds modeled on the complex numbers $\C$, which provide us with a wider operational framework and closed-form expressions.

%% file: SymmSpaces.tex

\section{Hermitian Symmetric Spaces}
We propose the systematic use of Hermitian symmetric spaces in representation learning. Symmetric spaces are Riemannian manifolds with rich symmetry groups. This makes them amenable to analysis and explicit computations. 
%
Hermitian symmetric spaces are modeled over the complex numbers $\C$ and carry an invariant complex structure. Furthermore, they admit matrix models of the Poincar\'e disc and the upper half-plane model of hyperbolic space in $\C^N$. This has several mathematical advantages, as it leads to a simple expression for the gradient of a function, and provides new analytical tools and a well-established theory of harmonic analysis. 

{\bf Matrix models:} We focus on the Siegel spaces $\mathcal{H}_n$ \cite{siegel1943symplectic} as examples of Hermitian symmetric spaces. Siegel spaces generalize the hyperbolic plane which is $\mathcal{H}_1$. They admit a bounded domain model and an upper half-space model in the space $\Sym(n,\C)$ of $n \times n$ symmetric matrices with complex entries. When $n=1$, these are the Poincar\'e model and the upper half-space model of the hyperbolic plane. The dimension of the space $\mathcal{H}_n$ is $n(n+1)$, and it contains $n$-dimensional Euclidean subspaces, products of $n$-copies of hyperbolic planes, as well as products of Euclidean and hyperbolic spaces.

A complex number $z\in\C$ can be written as $z=x+iy$ where $x,y\in\R$ and $i^2=-1$. Analogously a complex symmetric matrix $Z\in \Sym(n,\C)$ can be written as $Z = X+ iY$, where $X=\Re(Z),Y=\Im(Z) \in \Sym(n,\R)$ are symmetric matrices with real entries. 
We denote by $Z^*=X-iY$ the complex conjugate matrix.  
For a real symmetric matrix $Y \in \Sym(n,\R)$ we write $Y>\!\!>0$ to indicate that $Y$ is positive definite. 

\begin{wrapfigure}{R}{0.39\linewidth}
	\vspace{-5mm}
	\centering
	\begin{tikzpicture}[scale=0.5]
	\fill [fill=blue!40!white] (0,0) circle [radius= 1cm];
    \draw [thick] (0,0) circle [radius= 1cm];
    \filldraw (0,0)  circle [radius= 1pt] node [above right]{$0$};    
    \node at (1.1,1.4) {$\calB_1$};
    \node at (3,.5) {Cayley};
    \node at (3,0) {$\leftrightsquigarrow$};
    
    \fill[fill=blue!40!white] (5,-1) rectangle (8,1);
    \draw [thick] (5,-1) to (8.5,-1);    
    \filldraw (6.5,0)  circle [radius= 1pt] node [above right]{$i$};    
    \node at (8.2,1.4) {$\calS_1$};
    \draw (6.5, -1) to (6.5, 1.5);
    \node at (5.75,1.5) {$\Im z$};
    \node at (8.5,-1.5) {$\Re z$};
    \node at (6.52, 1.25) [rotate=90] {$>$};
    \node at (8.25, -1.05) {$>$};
	\end{tikzpicture}
	\caption{Matrix models.}
	\label{fig:spaces}
	\vspace{2mm}
\end{wrapfigure}

The bounded domain model is 
$$\calB_n:=\{Z\in\Sym(n,\C)|\;\Id-Z^*Z>\!\!>0\};$$
the Siegel space is 
$$\calS_n:=\{Z= X+iY\in\Sym(n,\C)|\;Y>\!\!>0\}. $$
An explicit isomorphism between the two domains given by the Cayley transform and explicit formulas for the distance are given in the Appendix \ref{a.models}.

%
%

{\bf Distance function:} We compute and algorithmically implement the distance function (see Algorithm~\ref{alg:distances}) using the Takagi factorization (see Appendix \ref{a.Takagi}) to obtain eigenvalues and eigenvectors of complex symmetric matrices in a tractable manner with automatic differentiation tools. 

\begin{wrapfigure}{R}{0.55\linewidth}
	\vspace{-7mm}
	\begin{minipage}{\linewidth}
		\begin{algorithm}[H]
			\caption{Algorithm to calculate the distance}
			\label{alg:distances}
			\begin{algorithmic}[1]
				\small
				\STATE Given two points $Z_1,Z_2\in \calS_n$
				\STATE Compute $Z_3:=\sqrt{\Im Z_1}^{-1}(Z_2-\Re Z_1)\sqrt{\Im Z_1}^{-1}\in\calS_n$
				\STATE Define $W=(Z_3 - i\Id) (Z_3 + i\Id)^{-1}\in\calB_n$
				\STATE Use the Takagi factorization to write
				$W=\ov K D K^*$ for $D$ real diagonal, and $K$ unitary.
				\STATE The distance is 
				$d^R(Z_1,Z_2):=\sqrt {\sum_{i=1}^n \left[\log{\frac{1 + d_i}{1 - d_i}}\right]^2}$, where $D={\rm diag}(d_i)$.
			\end{algorithmic}
		\end{algorithm}
	\end{minipage}%
\end{wrapfigure}
%
%
%
%

{\bf Optimization:} With the proposed matrix manifolds, we aim to apply Riemannian optimization \cite{bonnabel2011rsgd} in order to minimize loss functions based on Riemannian distances in the embedding space. We constrain the embeddings to remain within the models with a projection described in Appendix~\ref{a.project}.

{\bf Gradient:} To apply Riemannian optimization, we need an expression for the Riemannian gradient. For a function $f: \calS_n \to \R$ at a point $Z = X+iY \in \calS_n \subset \Sym(n,\C)$, it is given by ${\rm grad_R}(f) = Y \cdot {\rm grad_E}(f) \cdot Y$, where ${\rm grad}_E(f)$ is the Euclidean gradient of $f$ (see Appendix \ref{a.gradient}).

%% file: graphs.tex
\newpage
\section{Benchmarking graphs}
In order to investigate the representation capabilities of different geometric spaces, synthetic graphs have been used \cite{cruceru20matrixGraph, gu2019lmixedCurvature}. So far the attention was mainly on graphs with pure geometric features, such as grids or trees, or their Cartesian products. 
We propose to expand the set of synthetic benchmarking graphs by rooted products of graphs. 

\begin{figure}[b]
	\vspace{-4mm}
	\centering
	\subfloat{\label{fig:prod-cart-treegrid}{\includegraphics[clip, width=.25\linewidth,keepaspectratio]{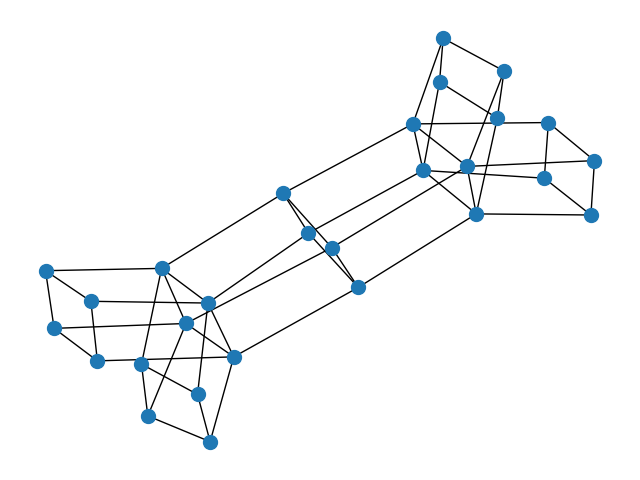}}}\hfill
	\subfloat{\label{fig:prod-cart-treetree}{\includegraphics[clip, width=.25\linewidth,keepaspectratio]{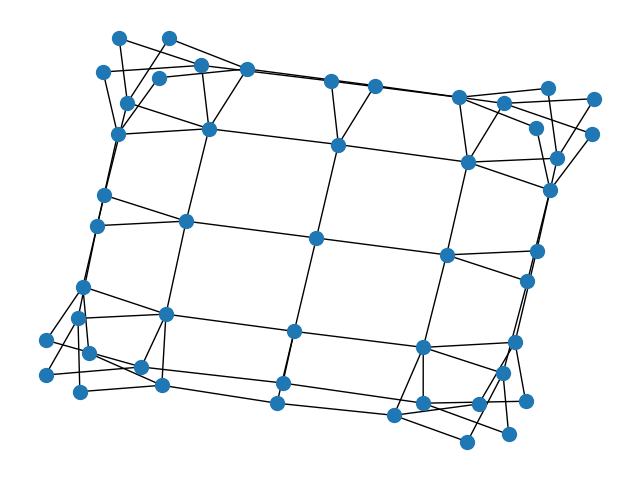}}}\hfill
	\subfloat{\label{fig:prod-root-treegrids}{\includegraphics[clip, width=.25\linewidth,keepaspectratio]{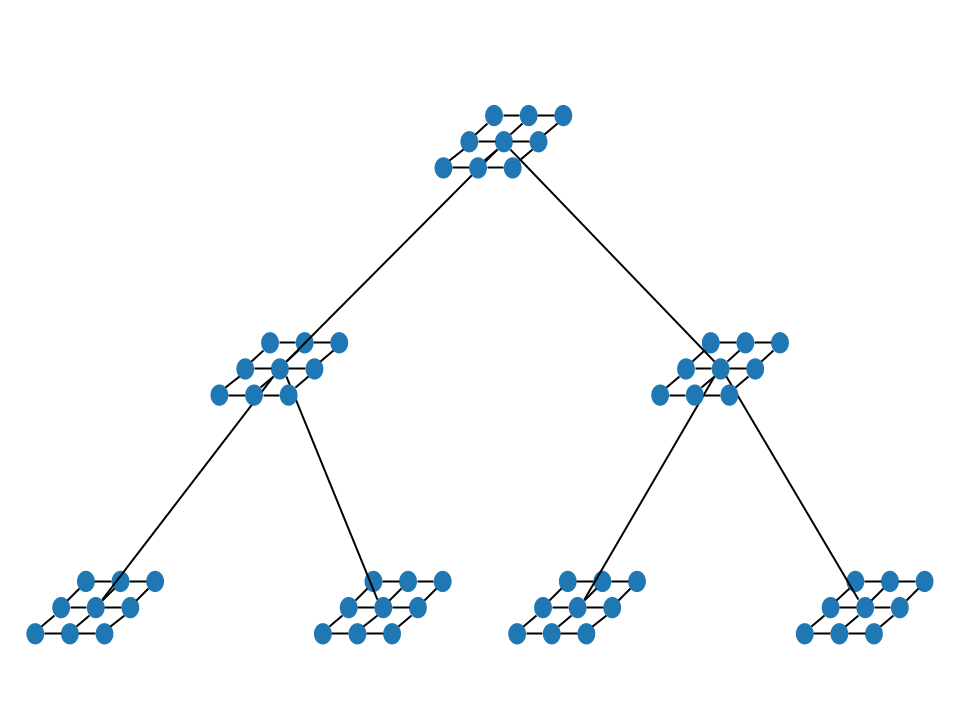}}}
	\subfloat{\label{fig:prod-root-gridtrees}{\includegraphics[clip, width=.25\linewidth,keepaspectratio]{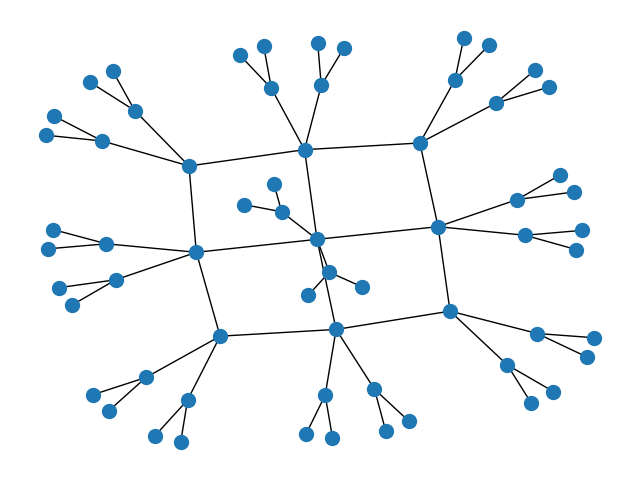}}}
	\caption{a) Cartesian product of tree and 2D grid. b) Cartesian product of tree and tree. c) Rooted product of tree and 2D grids. d) Rooted product of 2D grid and trees.}
	\label{fig:synthetic-graphs}
	\vspace{-2mm}
\end{figure}

Consider knowledge graphs as a motivating example. The same entity can exhibit multiple local logical patterns given by hierarchical, symmetric and anti-symmetric relations with other entities \cite{chami2020lowdimkge}. A similar case can be found in WordNet \cite{miller1995wordnet}. Words inside a synset show a full-mesh layout with synonyms, and tree-like hypernym relations with other synsets. As illustrated in Figure~\ref{fig:synthetic-graphs}, Cartesian products of trees and grids mix the tree- and grid-like features globally. On the other hand, rooted products of trees with grids and of grids with trees mix these features at different levels and scales. 
Thus, these graphs reflect to a greater extent the complexity of intertwining and varying structure in different regions, making them a better approximation of real-world datasets. Here we consider the rooted product \textsc{Tree $\diamond$ Grids} of a tree and  2D grids as well as the rooted product \textsc{Grid $\diamond$ Trees} of a 2D grid and trees. 
Iterating the procedure of rooted products, taking for example the rooted products with different graphs, or a rooted product of a rooted product, one can generate graphs with diverse structures at varying scales. 
%
%

%% file: Experiments.tex

\section{Experiments}

We evaluate the representation capabilities of the proposed approach for graph reconstruction. Following previous work \cite{deSa18tradeoffs, gu2019lmixedCurvature}, we measure the quality of the learned embeddings by reporting \textit{average distortion} $D_{avg}$ (global metric, lower is better) and \textit{mean average precision} (mAP, local metric, higher is better) as fidelity measures (formulas in Appendix~\ref{sec:appendix-metrics}).

\textbf{Baselines:} We compare our approach to constant-curvature baselines, such as Euclidean ($\mathbb{E}$) and hyperbolic ($\mathbb{H}$) spaces (we compare to the Poincar\'e model since the Bounded Domain model is a generalization of it), and Cartesian products thereof. To establish a fair comparison, we allow the same number of free parameters in each model. This is, the spaces $\calS_n$ and $\calB_n$ have $n (n + 1)$ parameters, thus we compare to baselines with the same dimensionality.

\textbf{Setup:} We experiment with the distortion loss proposed in \citet{gu2019lmixedCurvature} (formulas in Appendix~\ref{sec:appendix-loss}), which minimizes the relation between the distance in the space, compared to the distance in the graph, and captures the average distortion. As input and evaluation data we take the shortest distance in the graph between every pair of connected nodes. Unlike previous work \cite{cruceru20matrixGraph, gu2019lmixedCurvature} we do not apply any  scaling, neither in the input graph distances nor in the distances calculated on the space. 
Since the point $\textbf{0}$ is not part of the Siegel upper-half space, we initialize the matrix embeddings by adding small symmetric perturbations to the matrix basepoint $i\Id$ (see Appendix~\ref{a.random}). For the Bounded model, we additionally map the points with the Cayley transform. More experimental details in Appendix~\ref{sec:appendix-expdetails}.

\textbf{Synthetic Graphs:}
We report the performance of models and baselines in Table~\ref{tab:synthetic-results}. We find that the Hermitian models perform on par with the matching geometric spaces, or with the best-fitting product of spaces across all graphs of pure geometry (grids, trees and Cartesian products thereof). While constant-curvature baselines only perform well when the structure of the data conforms to their geometry, the matrix models are able to automatically adapt to very dissimilar patterns without any apriori estimates of the curvature or other features of the graph. This showcases the flexibility of our model, due to its enhanced geometry and higher expressivity.
Results for more dimensionalities in Appendix~\ref{sec:appendix-results}.

For graphs with mixed geometric features, such as the rooted products, the matrix models outperform all baselines. Cartesian products of spaces cannot arrange these compound geometries into separate Euclidean and hyperbolic subspaces. Hermitian symmetric spaces, on the other hand, offer a less distorted representation of these tangled patterns by exploiting their richer geometry which mixes hyperbolic and Euclidean features.  Moreover, they reach a competitive performance on the local neighborhood reconstruction, as the mean precision shows.

\begin{table}[!t]
	\caption{Results for synthetic datasets. All models have same number of free parameters.}
	\label{tab:synthetic-results}
	\centering
	\adjustbox{max width=\textwidth}{
		\begin{tabular}{cc@{\hskip 2.2mm}cc@{\hskip 2.2mm}cc@{\hskip 1.5mm}cc@{\hskip 1.5mm}cc@{\hskip 1.5mm}cc@{\hskip 1.5mm}c}
			\toprule
			& \multicolumn{2}{c}{\textsc{4D Grid}} & \multicolumn{2}{c}{\textsc{Tree}} & \multicolumn{2}{c}{\textsc{Tree $\times$ Grid}} & \multicolumn{2}{c}{\textsc{Tree $\times$ Tree}} & \multicolumn{2}{c}{\textsc{Tree $\diamond$ Grids}} & \multicolumn{2}{c}{\textsc{Grid $\diamond$ Trees}} \\
			$(|V|, |E|)$ & \multicolumn{2}{c}{$(256, 768)$} & \multicolumn{2}{c}{$(364, 363)$} & \multicolumn{2}{c}{$(135, 306)$} & \multicolumn{2}{c}{$(225, 420)$} & \multicolumn{2}{c}{$(496, 774)$} & \multicolumn{2}{c}{$(567, 570)$} \\
			\textbf{} & $D_{avg}$ & mAP & $D_{avg}$ & mAP & $D_{avg}$ & mAP & $D_{avg}$ & mAP & $D_{avg}$ & mAP & $D_{avg}$ & mAP \\
			\cmidrule(lr){2-13}
			$\mathbb{E}^{20}$ & \textbf{0.125} & 1.000 & 0.039 & 0.456 & 0.121 & 1.000 & 0.098 & 0.960 & 0.039 & 0.453 & 0.046 & 0.369 \\
			$\mathbb{H}^{20}$ & 0.244 & 0.811 & \textbf{0.005} & 1.000 & 0.176 & 0.926 & 0.207 & 0.736 & 0.119 & 0.528 & 0.118 & 0.324 \\
			$\mathbb{E}^{10} \times \mathbb{H}^{10}$ & \textbf{0.125} & 1.000 & 0.013 & 1.000 & 0.115 & 1.000 & 0.093 & 0.991 & 0.020 & 0.592 & 0.017 & 0.988 \\
			$\mathbb{H}^{10} \times \mathbb{H}^{10}$ & 0.186 & 0.917 & 0.006 & 1.000 & 0.127 & 1.000 & \textbf{0.086} & 0.976 & 0.018 & 0.695 & 0.021 & 0.924 \\
			$\mathcal S_4$ & \textbf{0.125} & 1.000 & 0.013 & 0.777 & \textbf{0.113} & 1.000 & 0.087 & 0.980 & \textbf{0.015} & 0.674 & \textbf{0.012} & 0.848 \\
			$\mathcal B_4$ & \textbf{0.125} & 1.000 & 0.013 & 0.781 & \textbf{0.113} & 1.000 & 0.087 & 0.979 & 0.017 & 0.700 & 0.013 & 0.974 \\
			\bottomrule
		\end{tabular}
	}
	\vspace{-3mm}
\end{table}

\begin{wraptable}{r}{0.43\linewidth}
	\vspace{-4.2mm}
	\caption{Results for real datasets.}
	\label{tab:real-results}
	\centering 
	\adjustbox{max width=0.98\linewidth}{ 
		\begin{tabular}{cccc}
			\toprule 
			& \multicolumn{2}{c}{\textbf{bio-diseasome}} & \textbf{USCA312} \\
			& $D_{avg}$ & mAP & $D_{avg}$ \\
			\cmidrule(lr){2-3}\cmidrule(lr){4-4}
			$\mathbb{E}^{20}$ & 0.0384 & 0.7603 & \textbf{0.0016} \\
			$\mathbb{H}^{20}$ & 0.0537 & 0.9194 & 0.0174 \\
			$\mathbb{H}^{10} \times \mathbb{E}^{10}$ & 0.0252 & 0.9104 & \textbf{0.0016} \\
			$\mathbb{H}^{10} \times \mathbb{H}^{10}$ & 0.0268 & 0.9334 & 0.0017 \\
			$\mathcal S_4$ & \textbf{0.0235} & 0.8628 & 0.0017 \\
			$\mathcal B_4$ & 0.0259 & 0.8904 & 0.0017 \\
			\bottomrule
		\end{tabular}
	}
\end{wraptable}

\textbf{Real Datasets:} We compare the models on the USCA312 dataset, of distances between North American cities, reported in previous work \cite{gu2019lmixedCurvature}, and on \textit{bio-diseasome}, a network of human disorders and diseases with reference to their genetic origins \cite{goh2007human}, on Table~\ref{tab:real-results}. 
On the USCA312 dataset, Hermitian symmetric spaces perform on par with the compared target manifolds. For the bio-diseasome dataset, the Siegel model outperforms all baselines. This shows the strong reconstruction capabilities of Hermitian symmetric spaces for real-world data as well. It also indicates that vertices in bio-diseasome form a network with a more intricate geometry, which the matrix model is able to unfold to a better extent. The possibilities of leveraging graph embeddings into Hermitian symmetric spaces for structural analysis of networks will be explored in future work.

%% file: Conclusions.tex

\section{Conclusions \& Future Work}

Riemannian manifold learning has regained attention due to appealing geometric properties that allow methods to represent non-Euclidean data arising in several domains \cite{rubindelanchy2020manifold}. 
We propose matrix models of Hermitian symmetric spaces, which are complex manifolds modelled on $\C$, to learn representations for undirected graphs. Our approach leverages the rich structure of these manifolds that combine aspects of Euclidean and hyperbolic geometry. We develop and implement tractable and mathematically sound algorithms to learn embeddings through gradient-descent methods.\footnote{Code available at: \url{https://github.com/fedelopez77/sympa}}
We suggest a new class of synthetic datasets for benchmarking graph embeddings. They are constructed as rooted products of grids and trees, exhibiting grid- and tree-like structures at different scales.
We showcase the effectiveness of the proposed approach on conventional as well as new datasets for the graph reconstruction task. 
Our method ties or outperforms constant-curvature and Cartesian product baselines without requiring any previous assumption on the geometric features. This shows the flexibility and enhanced representation capacity of Hermitian symmetric spaces.

As future directions, we consider extending the distance computation algorithms to Finsler metrics, developing the missing components for other optimization techniques, such as Riemannian Adam \cite{becigneul2019riemannianMethods}, further experimentation and analysis of the current models, and the application of the proposed techniques to a structural analysis of graphs as well as to downstream tasks.

%% file: appendix.tex
\section{Explicit Formulas for Siegel Spaces}

\subsection{Linear Algebra Conventions}
A few clarifications from linear algebra can be useful:
\begin{enumerate}
\item The inverse of a matrix $X^{-1}$, the product of two matrices $XY$, the square $X^2$ of a square matrix are understood with respect to the matrix operations. Unless all matrices are diagonal these are different than doing the same operation to each entry of the matrix.
\item If $Z=X+iY$ is a complex matrix, 
\begin{itemize}
\item$Z^t$ denotes the transpose matrix, i.e.  $(Z^t)_{ij}=Z_{ji}$, 
\item $\ov Z=X-iY$ denotes the complex conjugate
\item $X^*$ denotes its transpose conjugate, i.e. $X^*=\ov{X^t}$.
\end{itemize}
\item A complex square matrix $Z$ is \emph{Hermitian} if $Z^*=Z$. In this case its eigenvalues are real and positive. It is \emph{unitary} if $Z^*=Z^{-1}$. In this case its eigenvalues are complex numbers of absolute value 1 (i.e. points in the unit circle).
\item If $X$ is a real symmetric, or complex Hermitian matrix, $X>\!\!>0$ means that $X$ is positive definite, equivalently all its eigenvalues are bigger than zero.
\end{enumerate}

\subsection{Takagi Factorization}\label{a.Takagi}
Given a complex symmetric matrix $A$, the Takagi factorization is an algorithm that computes a real diagonal matrix $D$ and a complex unitary matrix $K$ such that 
$$A=\ov K D K^*.$$
This will be useful to work with the bounded domain model. It is done in a few steps
\begin{enumerate}
	\item Find $Z_1$ unitary, $D$ real diagonal such that
	$$A^*A=Z_1^* D^2 Z_1$$
	\item Find $Z_2$ orthogonal, $B$ complex diagonal such that 
	$$\ov Z_1AZ_1^*=Z_2BZ_2^t$$
	This is possible since the real and imaginary parts of $\ov Z_1AZ_1^*$ are symmetric and commute, and are therefore diagonalizable in the same orthogonal basis.
	\item Set $Z_3$ be the diagonal matrix with entries
	$$(Z_3)_{ii}=\left(\sqrt\frac{b_i}{|b_i|}\right)^{-1}$$
	where $b_i=(B)_{ii}$
	\item Set $K=Z_1^*Z_2Z_3$, $D$ as in Step 1. It then holds 
	$$A=\ov K D K^*.$$
\end{enumerate}

\subsection{Siegel Space and its Models}\label{a.models}
We consider two models for the symmetric space, the bounded domain
$$\calB_n:=\{Z\in\Sym(n,\C)|\;\Id-Z^*Z>\!\!>0\}$$
and the Siegel domain
$$\calS_n:=\{X+iY\in\Sym(n,\C)|\;Y>\!\!>0\}.$$
An explicit isomorphism between the two domains is given by the Cayley transform
$$\begin{array}{cccc}
c:&\calB_n&\to&\calS_n\\
&Z&\mapsto&i(Z+\Id)(Z-\Id)^{-1}
\end{array}$$ 
whose inverse $c^{-1}=t$ is given by
$$\begin{array}{cccc}
t:&\calS_n&\to&\calB_n\\
&X&\mapsto&(X-i\Id)(X+i\Id)^{-1}
\end{array}$$
The group of symmetries of the Siegel space $\calS_n$ is  
$\Sp(2n,\R)$, the subgroup of $\SL(2n,\R)$ preserving a symplectic form: a non-degenerate antisymmetric bilinear form on $\R^{2n}$. In this text we will choose the symplectic form represented, with respect to the standard basis, by the matrix $\bsm0&\Id_n\\-\Id_n&0\esm$ so that the symplectic group is given by the matrices that have the block expression
$$\left\{\bpm A&B\\C&D\epm\left|\begin{array}{l} A^tD-C^tB=\Id\\ A^tC=C^tA\\B^tD=D^tB\end{array}\right.\right\}$$
where $A,B,C,D$ are real $n\times n$ matrices.

The symplectic group $\Sp(2n,\R)$ acts on $\calS_n$ by non-commutative fractional linear transformations
$$\begin{pmatrix}A&B\\C&D\end{pmatrix}\cdot Z=(AZ+B)(CZ+D)^{-1}.$$
The action of $\Sp(2n,\R)$ on $\calB_n$ can be obtained through the Cayley transform.

\subsection{Crossratio and Distance}\label{a.cr}

{\bf \noindent Siegel upperhalf space:}
Given two points $X,Y\in\calS_n$ their crossratio is given by the complex $n\times n$-matrix 
$$R_\calS(X,Y)=(X-Y)(X-\ov Y)^{-1}(\ov X-\ov Y)(\ov X-Y)^{-1}.$$

It was established by Siegel \cite{siegel1943symplectic} 
that if $r_1,\ldots, r_n$ denote the eigenvalues of $R$ (which  are necessarily real greater than or equal to 1)
and we denote by $d_i$ the numbers
$$d_i=\log{\frac{1 + \sqrt r_i}{1-\sqrt r_i}}$$
Then the Riemannian distance  is given by
$$d^R(X,Y)=\sqrt {\sum_{i=1}^n d_i^2}.$$
Useful Finsler distances are given by
$$d^{F_1}(X,Y)=\sum_{i=1}^n d_i \qquad d^{F_\infty}(X,Y)=d_1.$$
There are explicit bounds between the distances, for example 
\begin{align}\label{e.dist}\frac 1{\sqrt n} d^{F_1}(X,Y)\leq d^R(X,Y)\leq d^{F_1}(X,Y)\end{align}
Furthermore, we have 
$$d^{F_1}(X,Y)=\log\det(\sqrt{R(X,Y)}+\Id)-\log\det(\Id-\sqrt{R(X,Y)})$$
which, in turn, allows to estimate the Riemannian distance using \eqref{e.dist}.
In general it is computationally difficult to compute the eigenvalues, or the squareroot, of a general complex matrix. However, we can use the determinant $\det_R$ of the matrix $R(X,Y)$ to give a lower bound on the distance:
$$\log{\frac{1 + \sqrt {\det_R}}{1-\sqrt {\det_R}}} \leq d^R(X,Y).$$

{\bf \noindent Bounded domain:}
The same study applies to pairs of points  $X,Y\in\calB$, but their crossratio should be replaced by the expression
$$R_\calB(X,Y)=(X-Y)(X-\ov {Y^{-1}})^{-1}(\ov{ X^{-1}}-\ov {Y^{-1}})(\ov{ X^{-1}}-Y)^{-1}.$$

\subsection{Computing Distances with the Takagi Factorization}\label{a.dist}
By means of the Takagi factorization we can define an alternative way to compute the distances. As always, we discuss the two models separately.

{\bf \noindent Siegel upperhalf space:}
Given as input two points $Z_1,Z_2\in \calS_n$ we do the following
\begin{enumerate}
\item Compute 
$$Z_3:=\sqrt{\Im Z_1}^{-1}(Z_2-\Re Z_1)\sqrt{\Im Z_1}^{-1}\in\calS_n$$
\item Define
$$W=t(Z_3)\in\calB$$
\item Use the Takagi factorization to write
$$W=\ov K D K^*$$ for some real diagonal matrix $D$ with eigenvalues between 0 and 1, and some unitary matrix $K$.
\item The distance is given by 
$$d^R(Z_1,Z_2):=\sqrt {\sum_{i=1}^n \left[\log{\frac{1+d_i}{1-d_i}}\right]^2}$$
Here, as before, $d_i$ denote the diagonal entries of the diagonal  matrix $D$.
\end{enumerate}

{\bf \noindent Bounded domain:}
In this case, given $W_1,W_2\in\calB$ we consider the pair $Z_1,Z_2\in\calS_n$ obtained applying the Cayley transform $Z_i=t (W_i)$. Then we can apply the previous algorithm, indeed 
$$d^R(W_1,W_2)=d^R(Z_1,Z_2).$$
\subsubsection{Riemannian Gradient}\label{a.gradient}
We consider on $\Sym(n,\C)$ the Euclidean metric given by 
$$\|V\|_E^2=\tr(V\ov V),$$
here $\tr$ denotes the trace, and, as above,  $V\ov V$ denotes the matrix product of the matrix $V$ and its conjugate. 

{\bf \noindent Siegel upperhalf space:}
The Riemannian metric at a point $Z \in \calS_n$, where $Z = X + iY$ is given by \cite{Sie} 
$$\|V\|_R^2=\tr(Y^{-1}VY^{-1}\ov V).$$
As a result we deduce that 
$${\rm grad}(f(Z)) = Y \cdot {\rm grad_E}(f(Z)) \cdot Y$$
{\bf \noindent Bounded domain:}
In this case we have 
$${\rm grad}(f(Z)) = A \cdot {\rm grad_E}(f(Z)) \cdot A$$
where $A = \Id - \overline{Z}Z$

\subsection{Embedding Initialization}\label{a.random}
Different embeddings methods initialize the points close to a fixed basepoint. In this manner, no a priori bias is introduced in the model, since all the embeddings start with similar values.

{\bf \noindent Siegel upperhalf space:} We choose as basepoint the matrix $i\Id$.

{\bf \noindent Bounded domain:} We choose the point \textbf{0}. 

In order to produce a random point we generate a random symmetric matrix with small entries and add it to our basepoint. As soon as all entries of the perturbation are smaller than $1/n$ the resulting matrix necessarily belongs to the model.
In our experiments, we generate random symmetric matrices with entries taken from a uniform distribution $\mathcal{U}(-0.001, 0.001)$.

\subsection{Projecting Back to the Models}\label{a.project}
The goal of this section is to explain two algorithms that, given $\epsilon$ and a point $Z\in\Sym(n,\C)$, return a point $Z_\epsilon^\calS$ (resp. $Z_\epsilon^\calB$), a point close to the original point lying in the $\epsilon$-interior of the model. This is the equivalent of the projection applied in \citet{nickel2017poincare} to constrain the embeddings to remain within the Poincar\'e ball, but adapted to the structure of the model. Observe that the projections are not conjugated through the Cayley transform.

{\bf \noindent Siegel upperhalf space:}
In the case of the Siegel upperhalf space $\calS_n$ given a point $Z=X+iY\in\Sym(n,\C)$
\begin{enumerate}
\item Find a real $n$-dimensional diagonal matrix $D$ and an orthogonal matrix $K$ such that $$Y=K^tDK$$
\item Compute the diagonal matrix $D_\epsilon$ with the property that
$$(D_\epsilon)_{ii}=\begin{cases} D_{ii} &\text{ if } D_{ii}>\epsilon\\\epsilon &\text{ otherwise}\end{cases}$$
\item The projection is given by
$$Z_\epsilon^\calS:=X+i K^tD_\epsilon K$$
\end{enumerate}

{\bf \noindent Bounded Domain:} In the case of the bounded domain $\calB$ given a point $Z=X+iY\in\Sym(n,\C)$
\begin{enumerate}
\item Use the Takagi factorization to find a real $n$-dimensional diagonal matrix $D$ and an unitary matrix $K$ such that $$Y=\ov KDK^*$$
\item Compute the diagonal matrix $D^\calB_\epsilon$ with the property that
$$(D_\epsilon^\calB)_{ii}=
\begin{cases} 
D_{ii} &\text{ if } D_{ii}<1-\epsilon\\
1-\epsilon &\text{ otherwise
}\end{cases}$$
\item The projection is given by
$$Z_\epsilon^\calB:=\ov KD^{\calB}_\epsilon K^*$$
\end{enumerate}

\section{Experimental Setup}
\label{sec:appendix-expdetails}

\subsection{Implementation of Complex Operations}

All models and experiments were implemented in PyTorch \cite{paszke2017pytorch}. Given a complex matrix $Z \in \C^{n \times n}$, we model real and imaginary components $Z = X + iY$ with $X, Y \in \R^{n \times n}$ separate matrices with real entries.

We followed standard complex math to implement basic arithmetic matrix operations. For complex matrix inversion we implemented the procedure detailed in \citet{falkenberg2007matrixinverse}.

\subsection{Optimization}
\label{sec:appendix-optimization}

As stated before, the models under consideration are Riemannian manifolds, therefore they can be optimized via stochastic Riemannian optimization methods such as \textsc{Rsgd} \cite{bonnabel2011rsgd} (we adapt the Geoopt implementation \cite{geoopt2019kochurov}).
Given a function $f(\theta)$ defined over the set of embeddings (parameters) $\theta$ and let $\nabla_{R}$ denote the Riemannian gradient of $f(\theta)$, the parameter update according to \textsc{Rsgd} is of the form:
$$\theta_{t + 1} = \mathcal{R}_{\theta_{t}}(-\eta_{t} \nabla_{R}f(\theta_{t}))$$

where $\mathcal{R}_{\theta_{t}}$ denotes the retraction onto space at $\theta$ and $\eta_{t}$ denotes the learning rate at time $t$. Hence, to apply this type of optimization we require the Riemannian gradient (described in Appendix~\ref{a.gradient}) and a suitable retraction.

\textbf{Retraction:}
Following \citet{nickel2017poincare} we experiment with a simple retraction:
$$\mathcal{R}_{\theta_{t}}(v) = \theta + v$$

\subsection{Loss Function}
\label{sec:appendix-loss}
To compute the embeddings, we optimize the distance-based loss function proposed in \citet{gu2019lmixedCurvature}. Given graph distances $\{d_{G}(X_i, X_j)\}_{ij}$ between all pairs of connected nodes, the loss is defined as:
$$\mathcal{L}(x) = \sum_{1 \leq i \leq j \leq n} \left| \left( \frac{d_{\mathcal{P}}(x_i, x_j)}{d_{G}(X_i, X_j)} \right)^2 - 1\right|$$

where $d_{\mathcal{P}}(x_i, x_j)$ is the distance between the corresponding node representations in the embeddings space.
This formulation of the loss function captures the average distortion.
We regard as future work experimenting with different distances in the space, or varying the loss function, similar to the ones proposed on \citet{cruceru20matrixGraph}.

\subsection{Evaluation Metrics}
\label{sec:appendix-metrics}

To measure the quality of the learned embeddings we follow the same fidelity metrics applied in \citet{gu2019lmixedCurvature}, which are distortion and precision.
The distortion of a pair of connected nodes $a, b$ in the graph $G$, where $f(a), f(b)$ are their respective embeddings in the space $\mathcal{P}$ is given by:
$$\operatorname{distortion}(a,b) = \frac{|d_{\mathcal{P}}(f(a), f(b)) - d_{G}(a, b)|}{d_{G}(a, b)}$$
The average distortion $D_{avg}$ is the average over all pairs of points. Distortion is a global metric that considers the explicit value of all distances.

The other metric that we consider is the mean average precision (mAP). It is a ranking-based measure for local neighborhoods that does not track explicit distances. Let $G = (V, E)$ be a graph and node $a \in V$ have neighborhood $\mathcal{N}_a = \{b_1, . . . , b_{\operatorname{deg}(a)}\}$, where $\operatorname{deg}(a)$ is the degree of $a$. In the embedding $f$, define $R_{a,b_i}$
to be the smallest ball around $f(a)$ that contains $b_i$ (that is, $R_{a,b_i}$ is the smallest set of nearest points required to
retrieve the $i$-th neighbor of $a$ in $f$). Then: 
$$\operatorname{mAP}(f) = \frac{1}{|V|} \sum_{a \in V} \frac{1}{\operatorname{deg(a)}} \sum_{i=1}^{|\mathcal{N}_a|} \frac{|\mathcal{N}_a \cap R_{a,b_i}|}{|R_{a,b_i}|}$$

\subsection{Data}

We employ NetworkX \cite{networkx2008hagberg} to generate the synthetic datasets, and their Cartesian and rooted products. The real-world datasets were downloaded from the Network Repository \cite{networkrepository2015rossi}. The statistics of all datasets reported in this work are presented in Table~\ref{tab:data-stat}. By triplets we mean the 3-tuple $(\textbf{u}, \textbf{v}, d(\textbf{u}, \textbf{v}))$, where $\textbf{u, v}$ represent connected nodes in the graph, and $d(\textbf{u,v})$ is the shortest distance between them.

\begin{table}[t]
	\caption{Datasets stats}
	\label{tab:data-stat}
	\centering
	\begin{tabular}{ccccccc}
		\toprule
		& \textbf{Nodes} & \textbf{Edges} & \textbf{Triplets} & \textbf{\begin{tabular}[c]{@{}c@{}}Grid \\ Layout\end{tabular}} & \textbf{\begin{tabular}[c]{@{}c@{}}Tree\\ Valency\end{tabular}} & \textbf{\begin{tabular}[c]{@{}c@{}}Tree\\ Height\end{tabular}} \\
		\midrule
		\textsc{3D Grid} & 125 & 300 & 7,750 & $5 \times 5 \times 5$ &  &  \\
		\textsc{4D Grid} & 256 & 768 & 32,640 & $(4)^4$ &  &  \\
		\textsc{Tree} & 364 & 363 & 66,066 &  & 3 & 5 \\
		\textsc{Tree $\times$ Grid} & 135 & 306 & 9,045 & $3 \times 3$ & 2 & 3 \\
		\textsc{Tree $\times$ Tree} & 225 & 420 & 25,200 &  & 2 & 3 \\
		\textsc{Tree $\diamond$ Grids} & 496 & 774 & 122,760 & $4 \times 4$ & 2 & 4 \\
		\textsc{Grid $\diamond$ Trees} & 567 & 570 & 160,461 & $3 \times 3$ & 2 & 5 \\
		bio-diseasome & 516 & 1,188 & 132,870 &  &  &  \\
		USCA312 & 312 & 48,516 & 48,516 &  &  & \\
		\bottomrule
	\end{tabular}
\end{table}

\subsection{Implementation Details}
For all models and datasets we run the same grid search and optimize the distortion loss, applying \textsc{Rsgd}. The implementation of all baselines are taken from Geoopt \cite{geoopt2019kochurov}. We train for $3000$ epochs, reducing the learning rate by a factor of $5$ if the model does not improve the performance after $50$ epochs, and early stopping based on the average distortion if the model does not improve after $150$ epochs. We use the burn-in strategy \cite{nickel2017poincare, cruceru20matrixGraph} training with a 10 times smaller learning rate for the first 10 epochs.
We experiment with learning rates from $\{0.02, 0.01, 0.005\}$, batch sizes from $\{128, 512\}$ and max gradient norm from $\{100, 300\}$.

\subsection{Experimental Observations}
We noticed that for some combinations of hyper-parameters and datasets, the learning process for the Bounded domain model becomes unstable. Points eventually fall outside of the space, and need to be projected in every epoch.
We did not observe this behavior on the Siegel model. We consider that these findings are in line with the ones reported on \citet{nickel2018lorentz}, where they observe that the Lorentz model, since it is unbounded, is more stable for gradient-based optimization than the Poincar\'e one.

\section{Further results}
\label{sec:appendix-results}

Here we present results for the same models, datasets and setups, for different dimensionalities.

\begin{table}[h]
	\vspace{-2mm}
	\caption{Results for synthetic datasets with $n = 2$. All models have same number of free parameters.}
	\centering
	\adjustbox{max width=\textwidth}{
	\begin{tabular}{ccccccccccccccc}
		\toprule
		& \multicolumn{2}{c}{\textsc{3D Grid}} & \multicolumn{2}{c}{\textsc{4D Grid}} & \multicolumn{2}{c}{\textsc{Tree}} & \multicolumn{2}{c}{\textsc{Tree $\times$ Grid}} & \multicolumn{2}{c}{\textsc{Tree $\times$ Tree}} & \multicolumn{2}{c}{\textsc{Tree $\diamond$ Grids}} & \multicolumn{2}{c}{\textsc{Grid $\diamond$ Trees}} \\
		& $D_{avg}$ & mAP & $D_{avg}$ & mAP & $D_{avg}$ & mAP & $D_{avg}$ & mAP & $D_{avg}$ & mAP & $D_{avg}$ & mAP & $D_{avg}$ & mAP \\
		\cmidrule(lr){2-3}\cmidrule(lr){4-5}\cmidrule(lr){6-7}\cmidrule(lr){8-9}\cmidrule(lr){10-11}\cmidrule(lr){12-13}\cmidrule(lr){14-15}
		$\mathbb{E}^{6}$ & \textbf{0.124} & 1.000 & \textbf{0.125} & 1.000 & 0.102 & 0.225 & 0.123 & 1.000 & 0.103 & 0.624 & 0.084 & 0.265 & 0.104 & 0.191 \\
		$\mathbb{H}^{6}$ & 0.249 & 0.928 & 0.244 & 0.811 & \textbf{0.011} & 1.000 & 0.176 & 0.925 & 0.204 & 0.736 & 0.130 & 0.530 & 0.132 & 0.339 \\
		$\mathbb{E}^{3} \times \mathbb{H}^{3}$ & \textbf{0.124} & 1.000 & \textbf{0.125} & 1.000 & 0.037 & 0.774 & \textbf{0.113} & 1.000 & 0.094 & 0.741 & 0.035 & 0.552 & 0.052 & 0.913 \\
		$\mathbb{H}^{3} \times \mathbb{H}^{3}$ & 0.183 & 0.945 & 0.186 & 0.920 & 0.021 & 0.763 & 0.128 & 1.000 & \textbf{0.086} & 0.976 & \textbf{0.027} & 0.587 & \textbf{0.028} & 0.801 \\
		$\mathcal S_2$ & 0.156 & 1.000 & 0.180 & 0.938 & 0.026 & 0.681 & 0.129 & 1.000 & 0.087 & 0.974 & 0.031 & 0.572 & 0.038 & 0.776 \\
		$\mathcal B_2$ & 0.156 & 1.000 & 0.179 & 0.947 & 0.025 & 0.674 & 0.129 & 1.000 & 0.087 & 0.975 & 0.034 & 0.582 & 0.037 & 0.869 \\
		\bottomrule
	\end{tabular}
}
\end{table}

\begin{table}[t]
	\vspace{-2mm}
	\caption{Results for synthetic datasets with $n = 3$. All models have same number of free parameters.}
	\centering
	\adjustbox{max width=\textwidth}{
	\begin{tabular}{ccccccccccccccc}
		\toprule
		& \multicolumn{2}{c}{\textsc{3D Grid}} & \multicolumn{2}{c}{\textsc{4D Grid}} & \multicolumn{2}{c}{\textsc{Tree}} & \multicolumn{2}{c}{\textsc{Tree $\times$ Grid}} & \multicolumn{2}{c}{\textsc{Tree $\times$ Tree}} & \multicolumn{2}{c}{\textsc{Tree $\diamond$ Grids}} & \multicolumn{2}{c}{\textsc{Grid $\diamond$ Trees}} \\
		& $D_{avg}$ & mAP & $D_{avg}$ & mAP & $D_{avg}$ & mAP & $D_{avg}$ & mAP & $D_{avg}$ & mAP & $D_{avg}$ & mAP & $D_{avg}$ & mAP \\
		\cmidrule(lr){2-3}\cmidrule(lr){4-5}\cmidrule(lr){6-7}\cmidrule(lr){8-9}\cmidrule(lr){10-11}\cmidrule(lr){12-13}\cmidrule(lr){14-15}
		$\mathbb{E}^{12}$ & \textbf{0.124} & 1.000 & \textbf{0.125} & 1.000 & 0.057 & 0.341 & 0.121 & 1.000 & 0.098 & 0.960 & 0.049 & 0.377 & 0.062 & 0.262 \\
		$\mathbb{H}^{12}$ & 0.248 & 0.921 & 0.244 & 0.811 & 0.018 & 0.870 & 0.176 & 0.925 & 0.207 & 0.751 & 0.128 & 0.533 & 0.117 & 0.338 \\
		$\mathbb{E}^{6} \times \mathbb{H}^{6}$ & \textbf{0.124} & 1.000 & \textbf{0.125} & 1.000 & 0.017 & 1.000 & 0.114 & 1.000 & 0.093 & 0.988 & 0.023 & 0.584 & 0.023 & 0.993 \\
		$\mathbb{H}^{6} \times \mathbb{H}^{6}$ & 0.182 & 0.967 & 0.186 & 0.916 & \textbf{0.009} & 0.995 & 0.127 & 1.000 & \textbf{0.086} & 0.976 & 0.019 & 0.670 & 0.018 & 0.916 \\
		$\mathcal S_3$ & \textbf{0.124} & 1.000 & 0.140 & 1.000 & 0.016 & 0.720 & 0.114 & 1.000 & 0.087 & 0.984 & \textbf{0.018} & 0.640 & \textbf{0.016} & 0.897 \\
		$\mathcal B_3$ & \textbf{0.124} & 1.000 & 0.140 & 1.000 & 0.016 & 0.787 & \textbf{0.113} & 1.000 & 0.087 & 0.983 & 0.019 & 0.662 & \textbf{0.016} & 0.942 \\
		\bottomrule
	\end{tabular}
}
\end{table}

\begin{table}[t]
\begin{minipage}{.5\linewidth}
	\caption{Results for real datasets with $n = 2$.}
	\centering 
	\begin{tabular}{cccc}
		\toprule
		\multicolumn{1}{l}{} & \multicolumn{2}{c}{\textbf{bio-diseasome}} & \textbf{USCA312} \\
		\multicolumn{1}{l}{} & $D_{avg}$ & \multicolumn{1}{l}{mAP} & $D_{avg}$ \\
		\cmidrule(lr){2-3}\cmidrule(lr){4-4}
		$\mathbb{E}^{6}$ & 0.0724 & 0.5492 & \textbf{0.0016} \\
		$\mathbb{H}^{6}$ & 0.0583 & 0.8415 & 0.0202 \\
		$\mathbb{E}^{3} \times \mathbb{H}^{3}$ & 0.0464 & 0.7426 & 0.0017 \\
		$\mathbb{H}^{3} \times \mathbb{H}^{3}$ & 0.0412 & 0.7567 & 0.0017 \\
		$\mathcal S_2$ & \textbf{0.0411} & 0.7419 & 0.0018 \\
		$\mathcal B_2$ & 0.0444 & 0.7578 & 0.0017 \\
		\bottomrule
	\end{tabular}
\end{minipage}%
\begin{minipage}{.5\linewidth}
	\caption{Results for real datasets with $n = 3$.}
	\centering 
	\begin{tabular}{cccc}
		\toprule
		\multicolumn{1}{l}{} & \multicolumn{2}{c}{\textbf{bio-diseasome}} & \textbf{USCA312} \\
		\multicolumn{1}{l}{} & $D_{avg}$ & \multicolumn{1}{l}{mAP} & $D_{avg}$ \\
		\cmidrule(lr){2-3}\cmidrule(lr){4-4}
		$\mathbb{E}^{12}$ & 0.0455 & 0.6993 & \textbf{0.0016} \\
		$\mathbb{H}^{12}$ & 0.0573 & 0.8975 & 0.0199 \\
		$\mathbb{E}^{6} \times \mathbb{H}^{6}$ & 0.0286 & 0.8741 & \textbf{0.0016} \\
		$\mathbb{H}^{6} \times \mathbb{H}^{6}$ & 0.0289 & 0.8900 & 0.0017 \\
		$\mathcal S_3$ & \textbf{0.0276} & 0.8288 & 0.0018 \\
		$\mathcal B_3$ & 0.0289 & 0.8298 & 0.0017 \\
		\bottomrule
	\end{tabular}
\end{minipage}
\end{table}